\documentclass[letterpaper, 10 pt, conference]{ieeeconf}  

\IEEEoverridecommandlockouts                              
\usepackage{cite}
\usepackage{amsmath,amssymb}
\usepackage{graphicx,mathtools,verbatim}
\usepackage[normalem]{ulem}
\usepackage{soul}
\usepackage{textcomp}
\graphicspath{{images/}}
\usepackage[dvipsnames]{xcolor}

\newcommand{\karen}[1]{\textcolor{red}{{Karen: #1}}}
\newcommand{\ck}[1]{\textcolor{Plum}{{Charlie: #1}}}

\newcommand{\alex}[1]{\textcolor{Bittersweet}{{Alex: #1}}}

\long\def\ignorethis#1{}

\newcommand{\etal}{{\em{et~al.}\ }}
\newcommand{\eg}{e.g.\ }
\newcommand{\ie}{i.e.\ }

\newcommand{\vc}[1]{\ensuremath{\mathbf{#1}}}

\newcommand{\argmin}{\operatornamewithlimits{argmin}}

\newcommand{\pctab}{\hspace{0.2in}}

\title{\LARGE \bf
Learning Human Behaviors for Robot-Assisted Dressing
}

\author{Alexander Clegg$^{1}$, Wenhao Yu$^{1}$, Jie Tan$^{2}$, Charlie C. Kemp$^{3}$, Greg Turk$^{1}$, C. Karen Liu$^{1}$ 
\thanks{$^{1}$School of Interactive Computing, Georgia Institute of Technology, Atlanta, GA., USA.}%
\thanks{$^{2}$Google Brain, Google, Mountain View, CA., USA.}%
\thanks{$^{3}$Department of Biomedical Engineering, Georgia Institute of Technology, Atlanta, GA., USA.}%
\thanks{This material is based upon work supported by the National Science Foundation Graduate Research Fellowship under Grant No. DGE-1650044 and NSF award IIS-1514258.}
}

\begin{document}

\maketitle
\thispagestyle{empty}
\pagestyle{empty}

\begin{abstract}

We investigate robotic assistants for dressing that can anticipate the motion of the person who is being helped. To this end, we use reinforcement learning to create models of human behavior during  assistance with dressing. To explore this kind of interaction, we assume that the robot presents an open sleeve of a hospital gown to a person, and that the person moves their arm into the sleeve. The controller that models the person\textquotesingle s behavior is given the position of the end of the sleeve and information about contact between the person\textquotesingle s hand and the fabric of the gown. We simulate this system with a human torso model that has realistic joint ranges, a simple robot gripper, and a physics-based cloth model for the gown.  Through reinforcement learning (specifically the TRPO algorithm) the system creates a model of human behavior that is capable of placing the arm into the sleeve. We aim to model what humans are \emph{capable of doing}, rather than what they \emph{typically do}. We demonstrate successfully trained human behaviors for three robot-assisted dressing strategies: 1) the robot gripper holds the sleeve motionless, 2) the gripper moves the sleeve linearly towards the person from the front, and 3) the gripper moves the sleeve linearly from the side.

\end{abstract}

\section{INTRODUCTION}\label{introduction}

The physical task of putting on an article of clothing can be difficult or impossible for people with limited mobility \cite{iom_2008,adls_1990,vest2011disability,soro2013_older}. While a number of specially designed devices exist to aid the dressing process, assistive robots might provide more adaptable and intelligent assistance to people in need. In many situations, robots could provide better assistance if they coordinated their motions with the person's motions. Ideally, the robot and the person would jointly optimize their motions to the person's benefit.  However, a brute force search of this space would be computationally intractable, and exploratory interactions between robots and real people can be onerous and risky. 

Research on robot-assisted dressing has relied on implicit and explicit models of human behavior, often making simplifying assumptions such as the human holding a fixed pose\cite{tamei2011reinforcement, chance2017quantitative} or the robot and the human interleaving their actions \cite{klee2015personalized}. Researchers have also considered data-driven approaches, such as methods that require human demonstrations \cite{shinohara2011learning, pignat2017learning} or body motion \cite{gao2015user}. 

Modeling human behavior is typically done by designing procedures to generate human motions based on prior knowledge or recorded motion data. Although this approach has been successful for reaching and locomotion~\cite{Yin:2007:SIM,Liu:2009:DMS,Ye:2010:OFC}, designing procedures to mimic what humans do for a loosely defined task, such as ``being dressed'', is likely to be biased to the specific data or assumptions. As a result, a robot controller developed to assist such human models is unlikely to take full advantage of the complementary capabilities of humans and robots. 

We propose a drastically different approach to human modeling. Instead of mimicking what humans \emph{typically do}, we aim to model what humans are \emph{capable of doing}. Specifically, our approach aims to answer the question: In the space of motions that a collaborative human is capable of performing, does a human motion that results in successful dressing exist for a given robotic system?

To this end, our goal is to develop a human control policy to complete the dressing task under a defined range of assistance provided by the robot, without using recorded human motion data or designing specific rules about ``being dressed''. Dressing is a challenging motor skill that requires utilizing multiple sensing modalities, such as vision and haptics, to manipulate highly deformable garments in a constrained space around the human body. We take the approach of reinforcement learning because the task of dressing is difficult to define by rules or structures and is highly sensitive to perceptual feedback.

While the recent advances in deep reinforcement learning (DRL) hold promise for learning complex motor skills, directly applying existing DRL techniques to the dressing problem is challenging for several reasons. First, the agent needs to learn to utilize haptic perception for two opposing tasks: \emph{applying force} to traverse inside of the garment and \emph{avoiding force} to prevent damage to the garment or itself. Furthermore, the haptic signals are only intermittently available to the agent depending on the occurrence of contact. Second, the only clear instruction for learning is the final goal, \ie, the agent must be dressed at the final state. Naively defining a delayed reward function based on the final goal might make the problem too difficult to learn. Lastly and most importantly, simulating rollouts for dressing scenarios is costly due to the cloth simulation of contact-rich scenes. This makes the rollout generation the limiting factor that significantly impacts the design of the reward function, states, and actions, rendering the end-to-end learning approach impractical. 

This paper is the first to demonstrate that it is possible to learn a robust control policy for human collaboration during robot-assisted dressing using reinforcement learning. We introduce a compact representation to encode haptics, vision, and knowledge about the garment as the input to the policy. This input space provides salient information for learning without overwhelming the learning algorithm under a strict rollout budget. While the input to the policy is restricted by the capability of human perception, the input to the reward function can take advantage of the simulation-based training framework that allows everything to be observable. As such, we design a metric to measure the progress of dressing that fully utilizes the state of both garment and human. Combining the progress metric with deformation and contact metrics, our reward function is able to give continuous feedback to differentiate very similar states during training.

We evaluate our approach on the scenario of assisting a person inserting an arm into a hospital gown. We show that the learned human policies can succeed at the dressing task under three types of robotic assistance: 1) the robot gripper holds the hospital gown in a fixed location, 2) the robot gripper moves linearly towards the person from the front, and 3) the gripper moves linearly from the side. We also show that our policy outperforms two baselines. The first baseline learns a policy without using the haptic information (Sections \ref{deformationpenalty} and \ref{haptics}) and the second one learns a policy without the task information(Section \ref{taskvector}).

\section{RELATED WORK}\label{relatedwork}

\subsection{User Modeling in Assistive Robot}

User behavior modeling has been used by robots to infer the state of a human user, such as the user's body pose or intentions. A common approach to model the user's behavior is to build a data-driven user preference model, which encodes the intention and possible actions of the user\cite{cakmak2011human, sisbot2007human, strabala2013towards}. For example, Cakmak \etal conducted a user study to learn the human preference on robot hand-over configurations. They combined the learned model with a kinematics-based planner to come up with hand-over configurations preferred by humans without loss of reachability\cite{cakmak2011human}. 

Due to occlusion of garments, it is difficult to collect user pose data during dressing events. Instead, existing work in robot-assisted dressing usually relies on prior knowledge about human behavior or constrained problem settings where estimation of user pose is feasible. For example, Yu \etal assume the user is holding a static pose while being dressed and used physics simulations for learning dressing outcome classifiers\cite{yu2017haptic}. A similar assumption was also made by Erickson \etal when training a regression model to estimate the pressure distribution on the user's arm during assistive dressing \cite{erickson2017how}. However, in general dressing scenarios this assumption is not likely to be true. Gao \etal used randomized decision forests to estimate the user pose from a top-view depth camera \cite{gao2015user}. They demonstrated their work with a Baxter robot that assisted a user with a sleeveless jacket, reducing the occlusion of the garment. Klee \etal proposed a method to optimize for not only the robot's motion, but also a request for user motion, which allowed them to dress a hat on the user\cite{klee2015personalized}. They used vision to monitor the user for completion of the requested movement, as well as to learn a mobility constraint model. However, when dressing the user with cloths or pants, vision-based monitoring becomes unreliable. 

\subsection{Deep Reinforcement Learning}

Deep reinforcement learning has been demonstrated on complex robotic motor skills with high dimensional state and action spaces\cite{lillicrap2015continuous,schulman2015high,schulman2015trust,mnih2016asynchronous}. These algorithms usually require a large amount of data to explore different regions of the state and action spaces, which makes rigid-body based control tasks its main application due to the efficiency of generating rollouts for these tasks. Directly applying deep reinforcement learning algorithms to learn motor skills in a deformable environment, such as assistive dressing, has previously been considered computationally infeasible. Tamei \etal applied reinforcement learning to dress a fixed mannequin with a t-shirt when the arms of the mannequin were already inside the sleeves\cite{tamei2011reinforcement}. They applied finite difference policy gradient to improve the control policy initialized using human demonstration. For more general dressing tasks, where a more capable control policy represented by a neural network model is desired, this approach is unlikely to succeed. Clegg \etal applied reinforcement learning to train a modular haptic feedback controller in a rigid body scenario and showed that it can be transferred directly to a deformable environment \cite{CleggYELT17}. They demonstrated self-dressing results using an aggregation of the learned controller with guiding trajectories for navigating inside the garment. In contrast, our method directly applies reinforcement learning in a dressing scenario, where the agent has to learn not only to avoid tearing the garment, but also to dress itself.

\section{Methods}\label{methods}

We develop a human control policy to complete the dressing task under a defined range of assistance that is provided by the robot, without using recorded human motion data or designing specific rules about ``being dressed''. We leverage deep reinforcement learning to provide an automatic, generic, and model-free approach to achieve this goal.

\subsection{Partially Observable Markov Decision Process}\label{mdp}
We formulate the assistive dressing task as a partially observable Markov Decision Process (POMDP) in order to learn a stochastic control policy $\pi: \mathcal{O} \times \mathcal{A} \mapsto [0, 1]$, which models the distribution of an action $\vc{a} \in \mathcal{A}$ conditioned on an observed state $\vc{o} \in \mathcal{O}$. A Markov Decision Process (MDP) is a tuple ($\mathcal{S}, \mathcal{A}, r, \rho, P_{sas'}, \gamma$), where $\mathcal{S}$ is the state space; $\mathcal{A}$ is the action space; $r$ is the reward function; $\rho$ is the distribution of the initial state $\vc{s}_0$; $P_{sas'}$ is the transition probability; and $\gamma$ is the discount factor. Our goal is to optimize the policy $\pi$, represented as a neural network, such that the expected accumulated reward is maximized.

Although in simulation we can access the full physical state of the human, the robot, and the garment, defined as $\vc{s} \in \mathcal{S}$, we formulate a MDP that is only \emph{partially observable} because humans do not have direct perception of the full state of the world and themselves. Giving our human model full observability might result in motions that real humans could not perform, and thus incorrectly suggest that a robot's policy can be helpful. As such, we design a compact observation space, $\mathcal{O}$, described in Section \ref{observationspace}. While the input to the policy, $\vc{o}$, should be restricted by the capability of human perception, the input to the reward function, $\vc{s}$, can take advantage of the full state of the simulated world. 
We propose a reward function (Section \ref{rewardfunction}) that quantifies dressing progress. 

\subsection{Reward Function}\label{rewardfunction}
The reward function quantifies the extent to which a state represents successful completion of a task. In this work, state $\vc{s}$ consists of the joint angles and velocities of the human model, the vertex positions of the garment, contact information from the previous simulation step and the values of a precomputed geodesic field on the garment. A good reward function is important to the success of reinforcement learning. Designing a reward function for the dressing task is nontrivial due to the difficulties of quantitatively defining dressing progress and balancing conflicting objectives, such as exploiting contact to push limbs through the garment while preventing large amounts of force that could tear apart the cloth. Additionally, care must be taken to ensure that reward is not too sparse for off-policy exploration to reach high reward states. For these reasons, we propose a novel reward function,
\begin{equation}
r(\mathbf{s})=w_1 \cdot r_{p}(\mathbf{s}) +w_2 \cdot r_{d}(\mathbf{s})+ w_3 \cdot r_{g}(\mathbf{s}) + r_{u}(\mathbf{s})
\end{equation}
where $ r_{p} $ is the progress reward, $ r_{d} $ is the deformation penalty, $ r_{g} $ is the geodesic reward, $w$ are the scalar weights of each term, and $ r_{u} $ encourages the torso to remain upright. In this work, we have empirically chosen $ w_1 = 5 $, $ w_2 = 6 $ and $w_3 = 2$ and we define each reward term in the following sections.

\begin{figure}[t]
\vspace{3mm}
\centering
\includegraphics[width=0.45\textwidth]{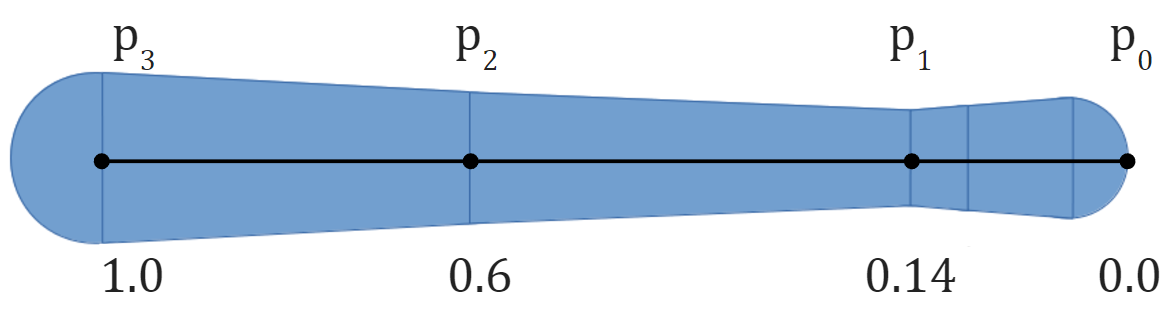}
\caption{Joint indices defined on our arm model (top) and values returned by the dressing progress metric when the garment feature contains the limb at various depths (bottom).} 
\label{fig:progress_metric}
\end{figure}

\begin{figure}[t]
\vspace{3mm}
\centering
\includegraphics[width=0.45\textwidth]{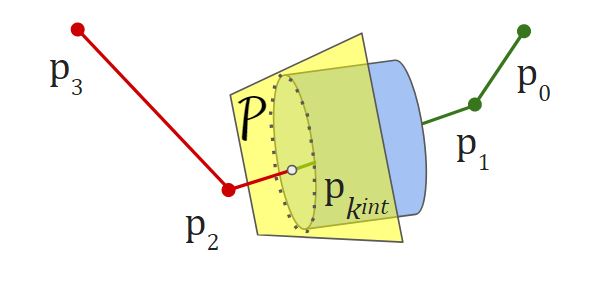}
\caption{Visualization of the containment computation for a limb partially inserted into a garment feature.} 
\label{fig:containment}
\end{figure}

\subsubsection{Progress Reward}\label{progressreward}
Without a continuous metric for progress, the dressing task becomes a sparse reward problem, making it challenging and data-inefficient for standard policy gradient methods. The progress reward metric measures the extend to which a limb is dressed at a state, $\mathbf{s}$. Figure~\ref{fig:progress_metric} shows joint indexing and progress measured along the limb being dressed in our experiments.

We first define a garment feature $\mathcal{F}$ as a set of indices to cloth vertices that best approximate a cloth structure of interest (e.g. the end of a sleeve or the waistband of a pair of pants)\cite{Clegg2015}. In our application, these features are often closed loops of vertices. For clarity of the exposition, we also define the world position of each joint of the limb as $\vc{p}_i$, where $i = 0, \cdots, m$. Unlike conventional joint index scheme, we index joints from distal to proximal. That is, $\vc{p}_m$ is the most proximal joint on the limb and $\vc{p}_1$ is the most distal. $\vc{p}_0$ is a dummy joint that refers the tip of the end-effector. All joint positions can be derived from the current state $\vc{s}$.

To measure progress, we first need to determine whether the intended limb is contained by the feature loop. If so, we reward the depth of the insertion. If not, we penalize the distance from the end-effector to the feature loop.

The containment of the limb in garment feature, $\mathcal{F}$, is computed as follows. First, we compute the best fit plane to the feature loop and project the feature vertices in $\mathcal{F}$ onto the plane, resulting in a 2D polygon $\mathcal{P}$. We then check whether a bone segment $\vc{b}_i$ between $\vc{p}_i$ and $\vc{p}_{i-1}$ intersects $\mathcal{P}$:
\begin{equation}
c_i(\mathbf{s}, \mathcal{F}) = \left\{
\begin{array}{ll} 1 & \mathrm{if\;\;} \vc{b}_i \cap \mathcal{P} \neq \emptyset \\
0 & \mathrm{otherwise.}
\end{array}
\right.
\end{equation}

Starting from the most distal bone segment, we check $c_i$ for each bone until the first encounter of $c_i = 1$. We then define $k_{int}$ as the index to the bone that intersects with the garment loop. If no bone intersects with the polygon, we set $k_{int} = 0$. The depth of containment is then defined as 
\begin{equation}
l(\vc{s}, \mathcal{F}) = | \vc{p}_{k_{int}-1} - \vc{r}| + \sum_{i=1}^{k_{int}-1} |\vc{b}_i|, 
\end{equation}
where $\vc{r} = \vc{b}_{k_{int}} \cap \mathcal{P}$. The reward function can be defined as: 

\begin{equation}
r_p(\vc{s}) = 
\left\{
\begin{array}{ll} l(\vc{s}, \mathcal{F}) & \mathrm{if\;\;} k_{int} \neq 0  \\
 -|\vc{c} - \vc{p}_0| & \mathrm{otherwise.}
\end{array}
\right.
\end{equation}
where $\vc{c}$ is the centroid of the polygon $\mathcal{P}$. 

\subsubsection{Deformation Penalty}\label{deformationpenalty}
In simulated dressing, the garment could be torn apart when it is deformed beyond reasonable strain limits by the movement of the human. The reward function penalizes these undesired states. Since we represent the cloth as a triangle mesh, we first define the deformation of a single cloth triangle, indexed by $i$, as the ratio of its current area, $a$, to its rest area, $a_{rest}$,
\begin{equation}
d_i(\vc{s}) = \frac{a}{a_{rest}}
\end{equation}

We then use the largest deformation across all triangles to calculate the deformation penalty. 
\begin{equation}
r_{d}(\mathbf{s}) = \tanh(w_{scale}(w_{thresh}-\max_i d_i(\vc{s})+2))-1
\end{equation}
where $w_{thresh}$ is a threshold defining the minimum deformation resulting in non-zero penalty. Since small deformations are a natural result of cloth dynamics, we only want to penalize deformations above this threshold. $w_{scale}$ scales the slope and upper limit of the deformation penalty function. We choose $w_{thresh}=15$ and $w_{scale}=0.7$ in all our experiments, effectively ignoring deformation below $15$ and capping the penalty at deformations exceeding $20$. In reality, humans exploit the deformation of the clothing when dressing some garments, and thus large deformation should not always be penalized heavily. For this reason, we choose to use tanh to cap the penalty. Note that an alternative approach to penalizing deformation would be to terminate any rollout in which the deformation is above a threshold. However, we observe that this strategy tends to punish exploration in the early stages of learning and results in overly-cautious policies.

\subsubsection{geodesic contact}\label{geodesiccontact}
When the limb is far from the garment feature, $\mathcal{F}$, $r_p(\mathbf{s})$ is the negative Cartesian distance to the loop centroid. This term alone is not enough to encourage successful policy behavior. The garment feature can be occluded by layers of cloth which must be moved aside in order to reach it and complete the dressing task. To guide the end effector through folds inside the cloth, we utilize the geodesic distance on the cloth to the garment feature. We first calculate a field $g(\vc{v})$ that maps a vertex $\vc{v}$ on the cloth mesh to the normalized geodesic distance between $\vc{v}$ and the garment feature (\ie vertices in $\mathcal{F}$ have distance $g(\vc{v}) = 0$ while those farthest from the feature have distance $ g(\vc{v}) = 1$). Among all the cloth vertices that are in contact with the end-effector $\mathcal{V}_c$, we select the one with the smallest geodesic distance:
\begin{equation}
g^*(\vc{s}) = \min_{\vc{v} \in \mathcal{V}_{c}(\vc{s})} g(\vc{v}). \end{equation}
such that if the end effector is touching the cloth, $ r_g(\mathbf{s}) = 1-g^*(\vc{s})$.

If the end effector is not touching the cloth, the reward is zero. This encourages contact with the garment and maximizes the number of haptic observations. Additionally, we only reward end effector contact with the cloth before the limb has entered the garment feature. Therefore, we return the maximum value if any part of the limb is contained by the feature. 

\begin{equation}
r_g(\mathbf{s}) = \left\{
\begin{array}{ll}
0 & \textrm{no contact,}\\
1 & \textrm{limb contained,}\\
1-g^*(\vc{s}) & \textrm{otherwise.}
\end{array}
\right.
\end{equation}

\ignorethis{
\subsubsection{Pose Target}\label{posetarget}\karen{I think we can just ignore this term and describe at beginning (when you describe $\vc{s}$) that the root is fixed in a position with limited orientation degree of freedom.}\alex{If you think that is the best way to handle this term, then we can remove this section.}
\ck{given the goal of exploring what is possible for the human to do, this term needs to be discussed with extra care. could it be described in terms of comfort or lack of motion from an initial pose? if only the torso is held upright, since all other weights are 0, maybe the role of this term can be described as being very narrow?}
The naturalness of the human pose is another important consideration in assistive dressing. Without a target pose, a trained policy may exhibit unnatural motions. Thus, we choose a natural target pose $\mathbf{q}_\textrm{target}$ and penalize poses that deviate from it:
\begin{equation}
\mathbf{r(s)_\textrm{pose}} = -||(\mathbf{q(s)} - \mathbf{q_\textrm{target}})*\mathbf{w_\textrm{pose}}||
\end{equation}
where $ \mathbf{q(s)} $ is the vector of joint angles describing the character pose at state $ \mathbf{s} $, $ \mathbf{q_\textrm{target}} $ is the pose prior, and $ \mathbf{w_\textrm{pose}} $ is a vector of weights describing the relative importance of maintaining each joint angle's pose prior. In our experiments, we only set the weight of the pelvis joint to be nonzero to encourage the character to maintain an upright torso posture. 
}

\subsection{Observation Space}\label{observationspace}

Some recent reinforcement learning approaches place the full state of the system into the observation. This approach is impractical in our dressing tasks, as the state dimension of the simulated cloth can be well beyond hundreds of thousands, which can result in an extremely large policy network that is infeasible to optimize. Additionally, we are interested in modeling what humans are capable of, and therefore we would like to limit policy input to the kind of information a human could reasonably be expected to have. For these reasons, we formulate a  compact observation space that is tailored for robot-assisted dressing tasks. The observation space includes the human's joint angles, garment feature locations, haptics, surface information and a task vector. 
\begin{equation}
\mathcal{O} = [\mathcal{O}_{p}, \mathcal{O}_f, \mathcal{O}_h, \mathcal{O}_s, \mathcal{O}_t]
\end{equation}
With carefully picked components, our observation is a 163-dimensional vector.

\subsubsection{Proprioception}\label{proprioception}
Following other recent work on reinforcement learning of motor control policies, we provide the policy with a proprioceptive input feature,
\begin{equation}
\mathcal{O}_p = [\cos(\mathbf{q(s)}), \sin(\mathbf{q(s)}), \mathbf{\dot{q}(s)}]
\end{equation}
where $ \mathbf{q(s)} $ is the vector of joint angles describing the human pose at state $ \mathbf{s} $. The human model in this work contains 22 degrees of freedom, 11 of which (excluding head/neck and left arm) are actuated. 

\subsubsection{Garment feature location}\label{garmentfeaturelocation}

The current location of a garment feature (\eg, a sleeve opening) and its relative position to the end effector are important information humans use for dressing, because the typical garment has more than one feature which could be dressed. We provide the policy with the world position of the centroid, $\vc{c}$, of the garment feature polygon, $\mathcal{P}$, and the displacement vector between the end effector and the centroid:
\begin{equation}
\mathcal{O}_f = [\mathbf{c}(\vc{s}), \vc{p}_0(\vc{s}) - \vc{c}(\vc{s})]
\end{equation}

\subsubsection{Haptics}\label{haptics}

\begin{figure}[t]
\vspace{3mm}
\centering
\includegraphics[width=0.35\textwidth]{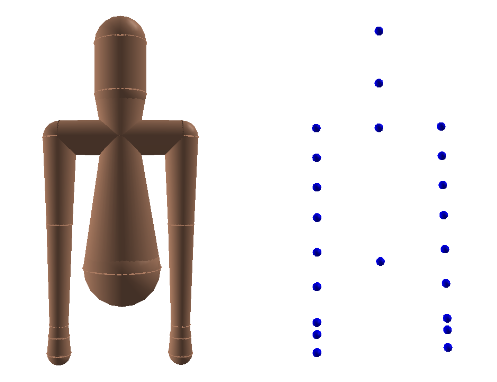}
\caption{Left: Human model used in this work. Right: Placement of haptic sensors on the human model.} 
\label{fig:humansensorsfigure}
\end{figure}

Humans rely on haptic sensing during dressing to avoid damage to clothes and to minimize discomfort. Inspired by Clegg \etal \cite{CleggYELT17}, we distribute a series of haptic sensors along the medial axis of the body nodes of the human character. We then construct a haptic feature vector by aggregating all contact forces between the human's body and the cloth. We do so by summing each contact into the closest haptic sensor, resulting in a haptic feature:

\begin{figure*}[t!]
\begin{tabular}{ccc}
  \includegraphics[width=55mm]{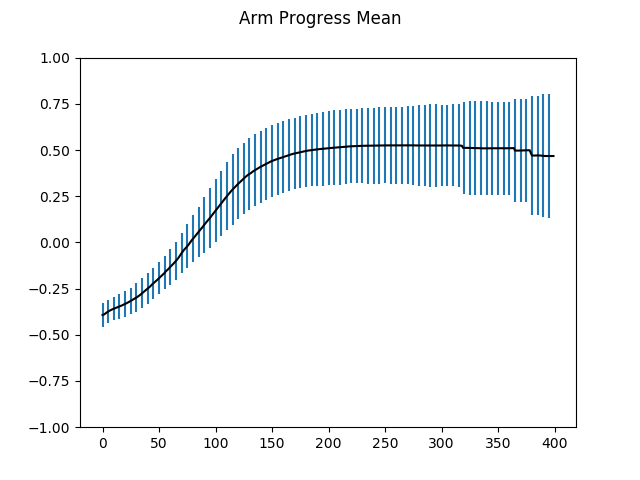} &
  \includegraphics[width=55mm]{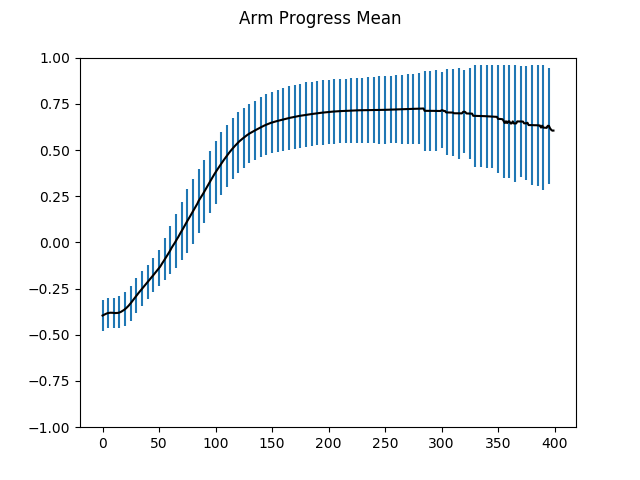} &
  \includegraphics[width=55mm]{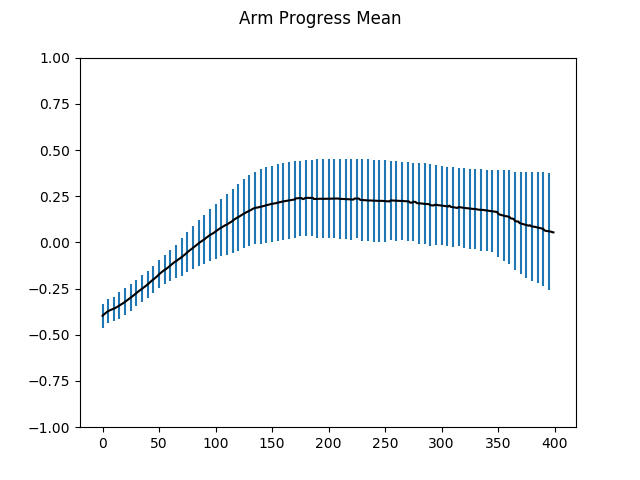} \\
 \includegraphics[width=55mm]{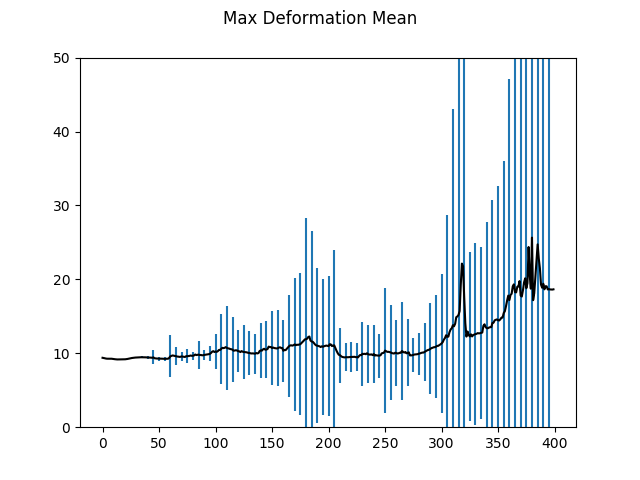} &  
 \includegraphics[width=55mm]{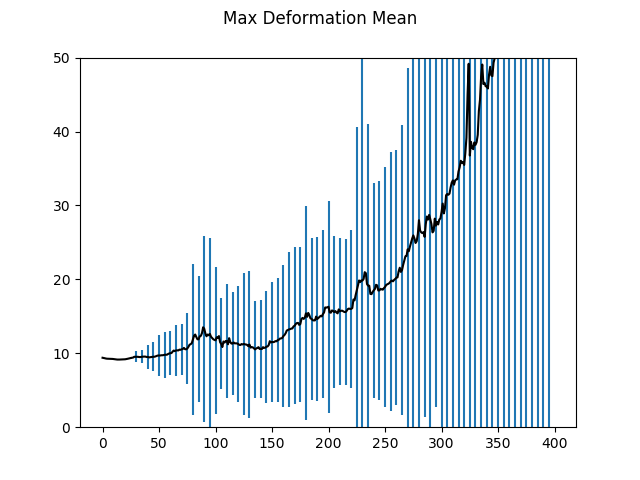} &
 \includegraphics[width=55mm]{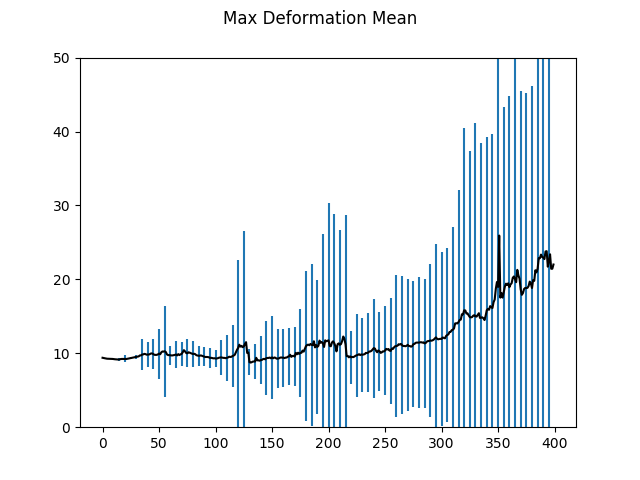} \\
(a) Complete & (b) No Haptics & (c) No Task \\[6pt]
\end{tabular}
\caption{Comparison of mean and standard deviation of Arm Progress and Max Deformation metrics for 100 random rollouts of the Fixed Gown task executed with our proposed combination of reward terms and observation features.}
\label{fig:baselines}
\end{figure*}

\begin{equation}
\mathcal{O}_h = [\mathbf{f}_0(\vc{s}), \cdots, \mathbf{f}_n(\vc{s})]
\end{equation}
where $ \mathbf{f}_i(\vc{s}) $ is the 3-dimensional accumulated force vector of the $ i $-th sensor and $ n = 21 $ for our human model, placed as shown in Figure \ref{fig:humansensorsfigure}. This haptic information encodes the location of contacts, as well as the magnitude and direction of the contact forces, which are essential to respecting the garment strain constraints.

\subsubsection{Signed surface} \label{contactids}

When putting on a shirt, humans typically stretch their arms through the sleeve while making contact with the inside of the garment. While humans seem to be able to distinguish the inner surface of a garment from the outer surface, this poses a great challenge for the simulated human with no vision and low-resolution tactile sensors. As such, we provide the policy with a surface sign for each haptic sensor $i$, that differentiates the contact between the inner and outer surfaces of the garment. We consider that in many common situations this information could be acquired by a combination of vision and haptic perception.
\begin{equation}
s_i = \left\{
\begin{array}{ll}
0 & \textrm{if not in contact,}\\
1 & \textrm{if inner surface contact,}\\
-1 & \textrm{if outer surface contact.}
\end{array}
\right.
\end{equation}
The surface sign for a single sensor is determined by comparing the direction of the contact force with the outward facing vertex normal for each contact point binned into that sensor in the current state. When these vectors are opposing, the cloth is being pushed from its outer surface, otherwise it is being pushed from the inner surface. The result is a surface sign feature,

\begin{equation}
\mathcal{O}_s = [s_0, \cdots, s_n],
\end{equation}
where $ n = 21 $ for our human model.

\subsubsection{Task vector} \label{taskvector}

Since our network architecture has no memory (no recurrency), the high level planning and order of events required to robustly complete the dressing task present significant challenges, making recovery from mistakes nearly impossible. For example, in the case that the end effector misses the garment on the first attempt, an optimal policy must reduce its immediate reward by backtracking in order to insert the limb into the feature, thus achieving higher total reward in a later state. 

To overcome these challenges, we provide the policy with a unit length task vector that suggests a direction for the end effector to move based on dressing progress. This task vector mimics a human's instinctual understanding of how to make progress on the dressing task based on prior knowledge about garment geometry and unique textural hints about garment features and the contact surface such as layers and seams which can not be obtained efficiently with existing cloth simulation techniques. This vector directs the end effector toward the garment feature until it makes contact with the garment and through the garment feature once it contains the limb. When the end effector is in contact with the garment but the limb is not yet contained, the task vector guides the end effector toward the feature along the gradient of the geodesic field defined on the garment. This allows the policy to backtrack from mistakes and to navigate the folds and occlusions between limb and feature, akin to the method that humans use to navigate a garment with touch alone. 

The task vector depends on geodesic information when the limb is in contact with the garment but has not yet entered the garment feature. We compute the gradient of the geodesic field $g(\vc{v})$ evaluated at $v^* = \argmin_{\vc{v} \in \mathcal{V}_{c}} g(\vc{v})$, the vertex that is in contact with the limb and is the closest to the garment feature:
\begin{equation}
\mathcal{O}_t = \left\{
\begin{array}{ll}
\frac{\vc{c} - \vc{p}_0}{||\vc{c} - \vc{p}_0||} & \textrm{if not in contact,}\\
\mathbf{n}_{plane} & \textrm{if limb contained,}\\
\nabla g(\vc{v})|_{\vc{v} = \vc{v}^*} & \textrm{otherwise.}
\end{array}
\right.
\end{equation}
where $ \mathbf{n}_{plane} $ is the normal of the feature plane.

\section{RESULTS}\label{results}
\begin{figure}[t]
\begin{tabular}{cc}
  \includegraphics[width=0.225\textwidth]{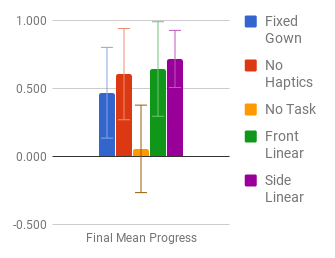} &
  \includegraphics[width=0.225\textwidth]{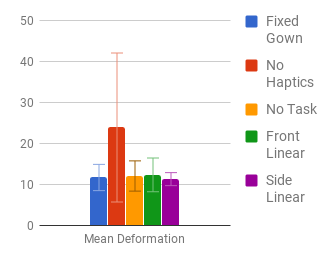} \\
\end{tabular}
\caption{Comparison graphs of mean and standard deviation for quantities from 100 random samples of each policy. The left bar graph shows the arm progress metric at the end of the rollout horizons, and the right shows the maximum deformation metric over the full durations of the rollouts.}
\label{fig:meanbars}
\end{figure}

\begin{figure}[t!]
\centering
\begin{tabular}{cc}
\includegraphics[width=0.225\textwidth]{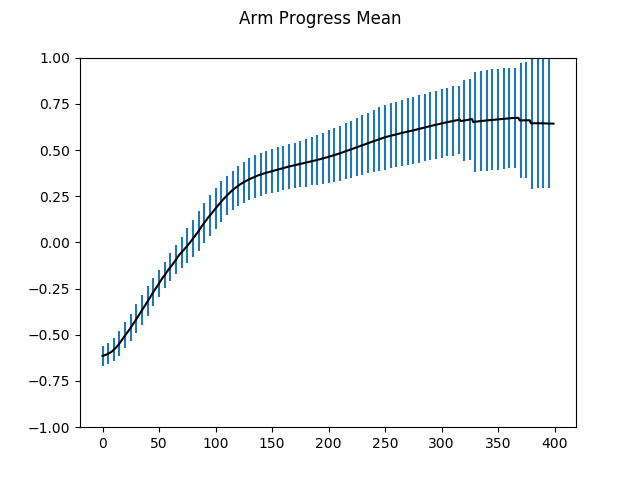} &
\includegraphics[width=0.225\textwidth]{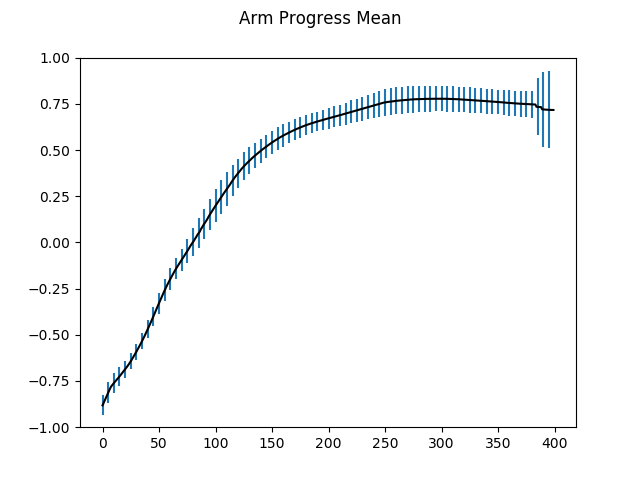} \\
\includegraphics[width=0.225\textwidth]{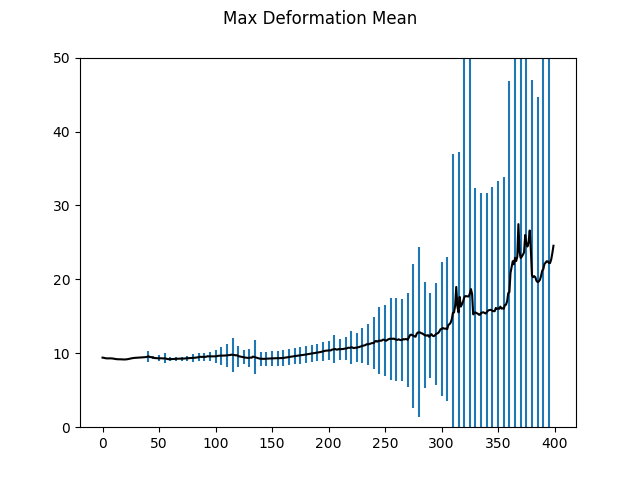} &
\includegraphics[width=0.225\textwidth]{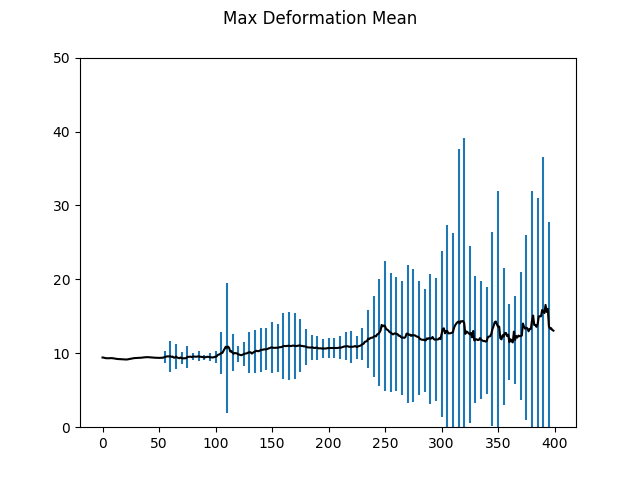} \\
Front Linear & Side Linear \\[6pt]
\end{tabular}
\caption{Mean and standard deviation of Arm Progress and Max Deformation metrics for 100 random rollouts of the Front Linear Gown  task and Side Linear Gown task.}
\label{fig:linearpolicies}
\end{figure}

The simulated dressing tasks in this work are implemented as environments in OpenAI's rllab reinforcement learning API \cite{Duan:2016:BDR:3045390.3045531}.  Character dynamics are simulated by DART \cite{DART}, and cloth dynamics by position based dynamics implemented by NVIDIA PhysX \cite{Macklin:2014, physx}. The hospital gown is represented by a triangle mesh and was created for use in prior work on modeling assistive dressing by Yu et. al \cite{yu2017haptic}. All policies are fully connected neural networks with 2 hidden layers of 64 nodes each and tanh activations. Policies were trained until reward reached a plateau, and required between [1000, 2000] TRPO iterations with a discount factor of 0.995, a frame-skip of 4 with simulation timestep 0.01, a 400 step rollout horizon and 50k sample steps per iteration. Our character model is displayed in Figure \ref{fig:humansensorsfigure} and consists of 22 degrees of freedom, 11 actuated by policy outputs. The root of the character model is fixed to the origin such that no translation is permitted.

To evaluate our proposed approach, we trained control policies for three simulated robot-assisted dressing strategies in which a gripper holds a hospital gown by the right sleeve and the actuated human model attempts to insert its right arm into the sleeve while the gripper is either motionless or moved along a linear trajectory. These tasks are described in more detail in the following sections.

\subsection{Fixed Gown}\label{fixedgown}
We initialize the garment with gripper position drawn uniformly from a rectangular prism with dimensions [0.6, 0.35, 0.1]. The gripper position remains fixed throughout the simulation. Despite its simplicity, this task provides the distinct challenge of forcing the character to move around the garment instead of relying on the garment to move into range.

\subsection{Front Linear Gown Trajectory}\label{frontlinear}
We initialize the garment with gripper position drawn uniformly from a rectangular prism with dimensions [0.4, 0.45, 0.1] approximately 0.6 meters in front of the character. A target location for the gripper is drawn from another rectangular prism with dimensions [0.3, 0.3, 0.1] approximately 0.2 meters in front of the character. The gripper moves the gown at a constant velocity between these two points over the course of 10 seconds and then remains fixed at the target location. This task simulates a naive dressing assistance policy whereby the robot moves the garment along a pre-computed linear trajectory toward the region of space initially occupied by the human's shoulder.

\subsection{Side Linear Gown Trajectory}\label{sidelinear}
We initialize the garment with gripper position drawn uniformly from a rectangular prism with dimensions [0.1, 0.3, 0.5] approximately 0.6 meters away from the character's right side. A target location for the gripper is drawn from another rectangular prism with dimensions [0.1, 0.3, 0.3] approximately 0.2 meters away from the character's right side. The gripper moves at a constant velocity between the two points over the course of 10 seconds and then remains fixed at the target location. For this task, the garment is rotated such that the garment feature plane normal (ie. the inner sleeve opening direction) aligns with the projection of the gripper trajectory on the XZ plane. This task simulates a naive linear dressing assistance policy to the side of the character where both the robot and the human may have a greater range of motion.

\subsection{Discussion}\label{discussion}
\begin{figure*}[t!]
\begin{tabular}{cccc}
\includegraphics[width=40mm]{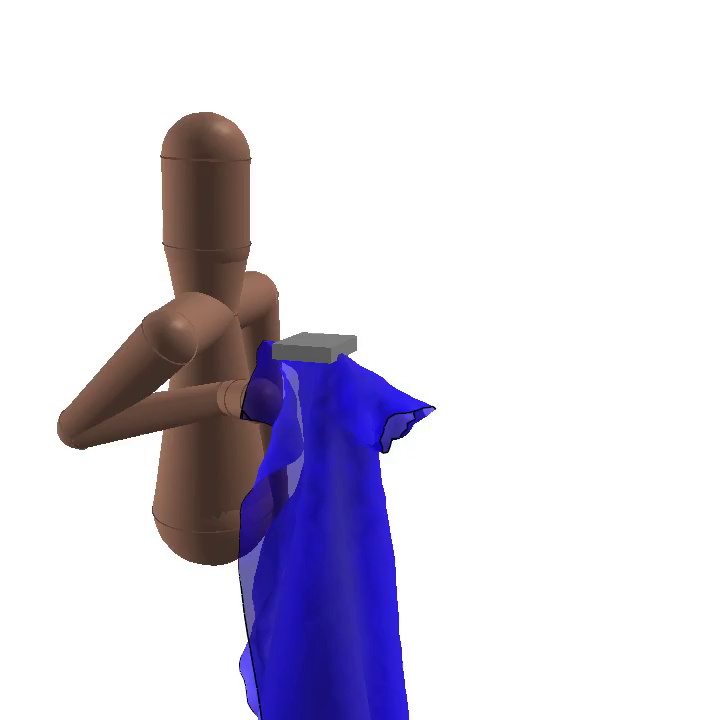} &
\includegraphics[width=40mm]{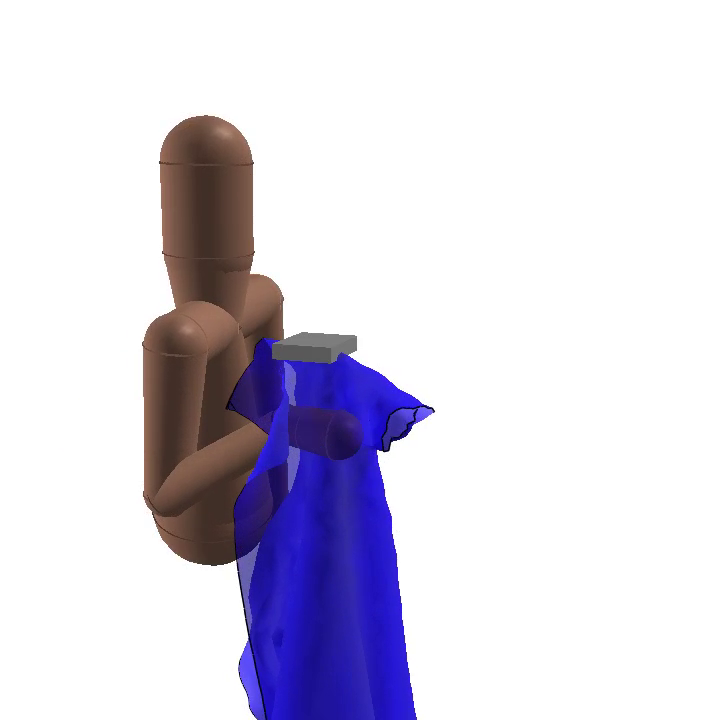} & 
\includegraphics[width=40mm]{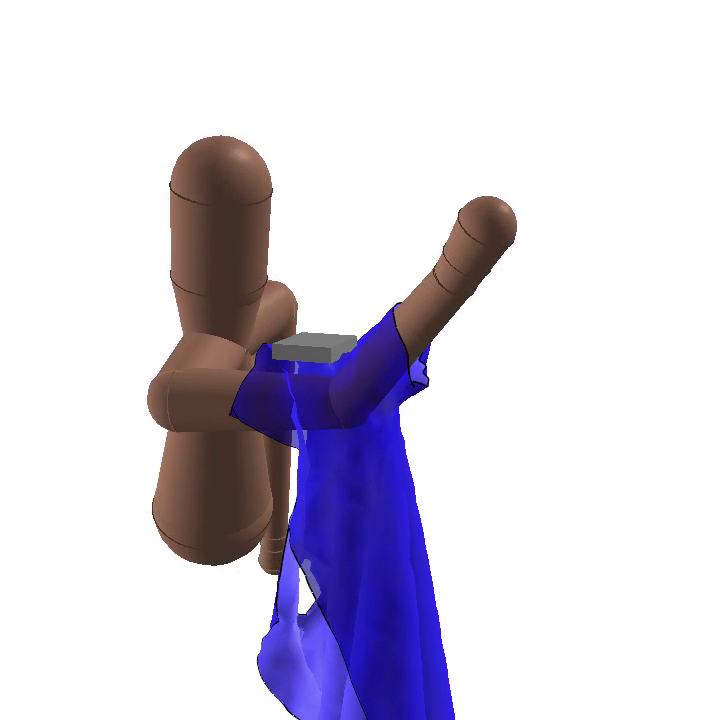} & 
\includegraphics[width=40mm]{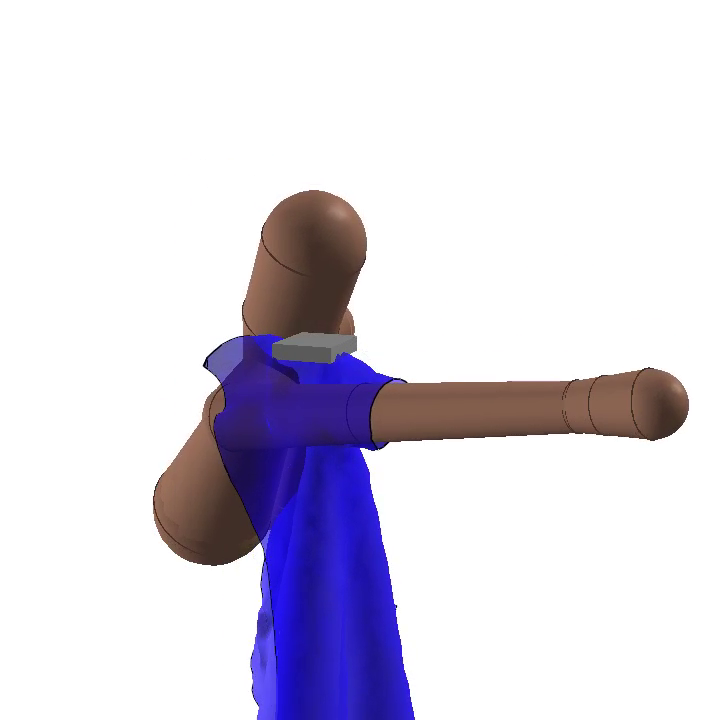} \\
\end{tabular}
\caption{Still frames of the our trained policy executing one rollout of the Fixed Gown task.}
\label{fig:mode7figure}
\end{figure*}

\begin{figure*}[t!]
\begin{tabular}{cccc}
\includegraphics[width=40mm]{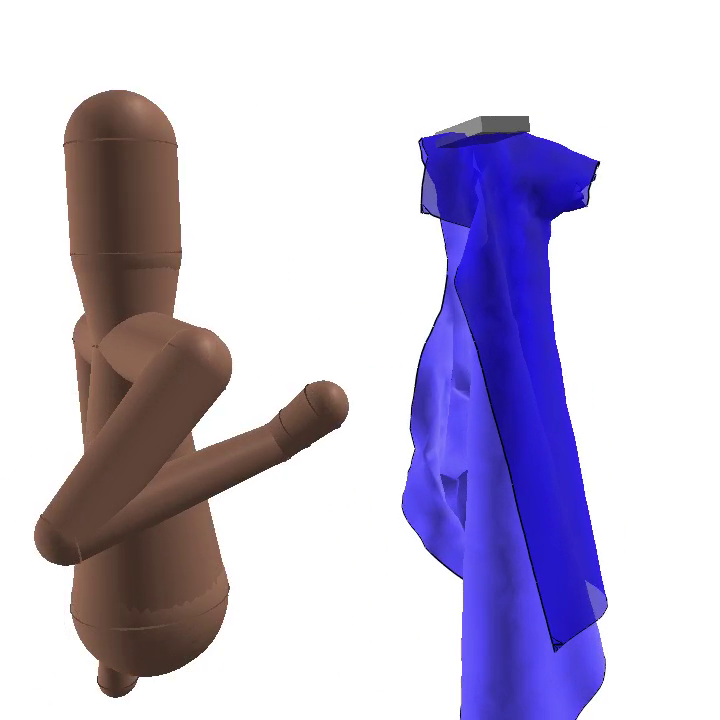} &
\includegraphics[width=40mm]{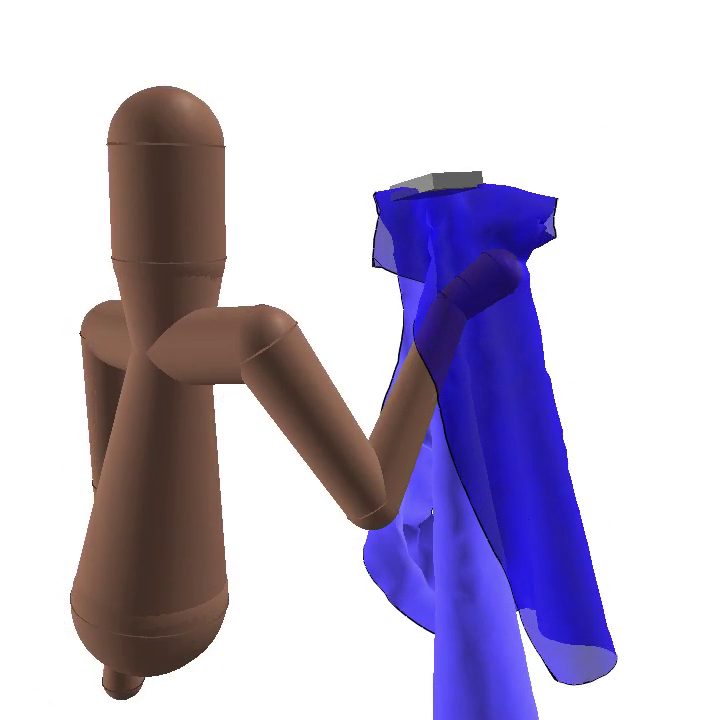} & 
\includegraphics[width=40mm]{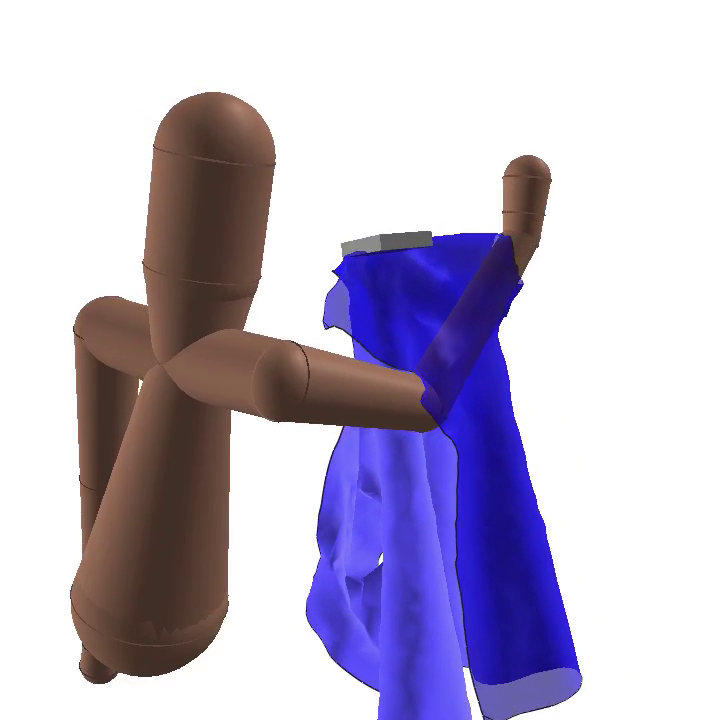} & 
\includegraphics[width=40mm]{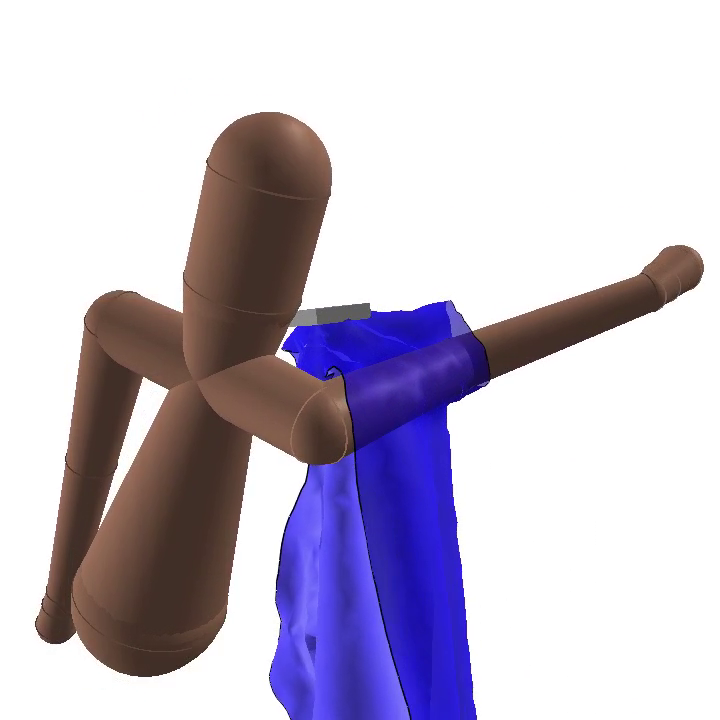} \\
\end{tabular}
\caption{Still frames of the our trained policy executing one rollout of the Front Linear Gown task.}
\label{fig:mode5figure}
\end{figure*}

\begin{figure*}[t!]
\begin{tabular}{cccc}
\includegraphics[width=40mm]{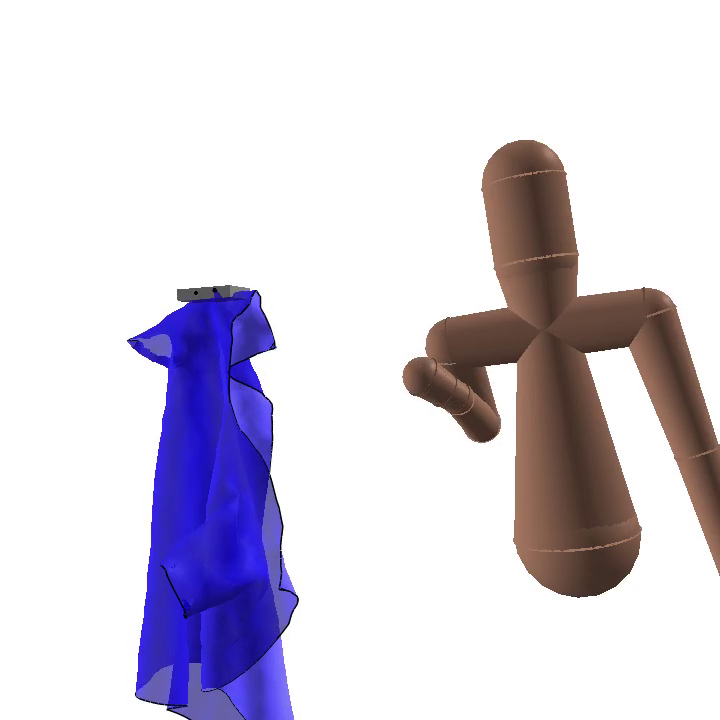} &
\includegraphics[width=40mm]{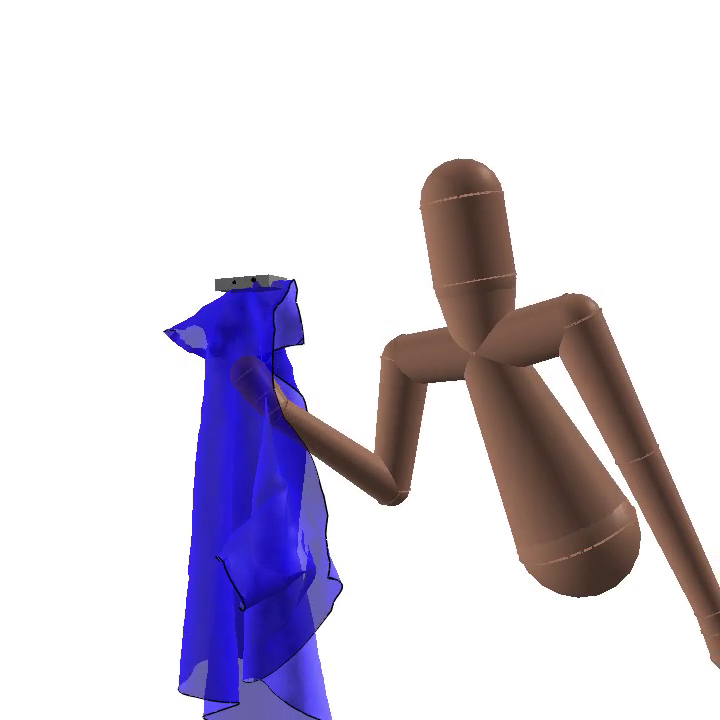} & 
\includegraphics[width=40mm]{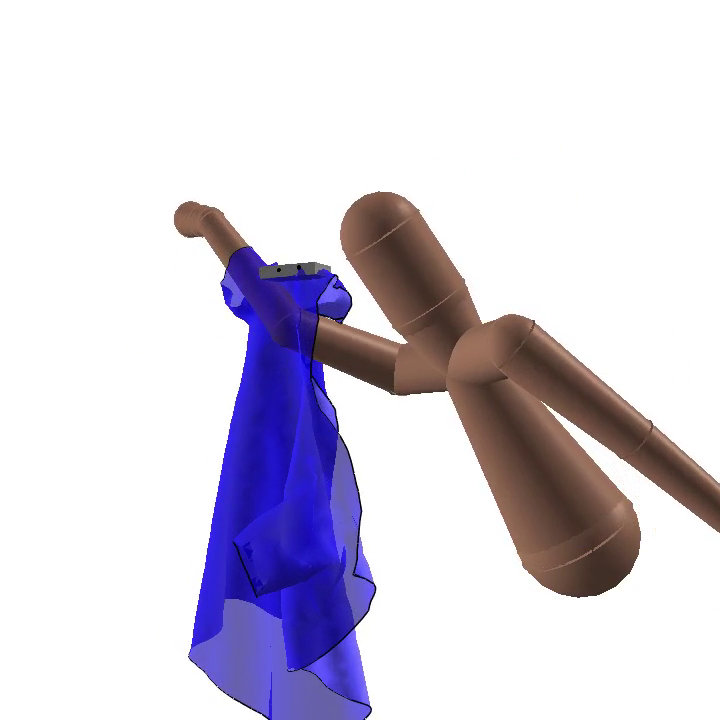} & 
\includegraphics[width=40mm]{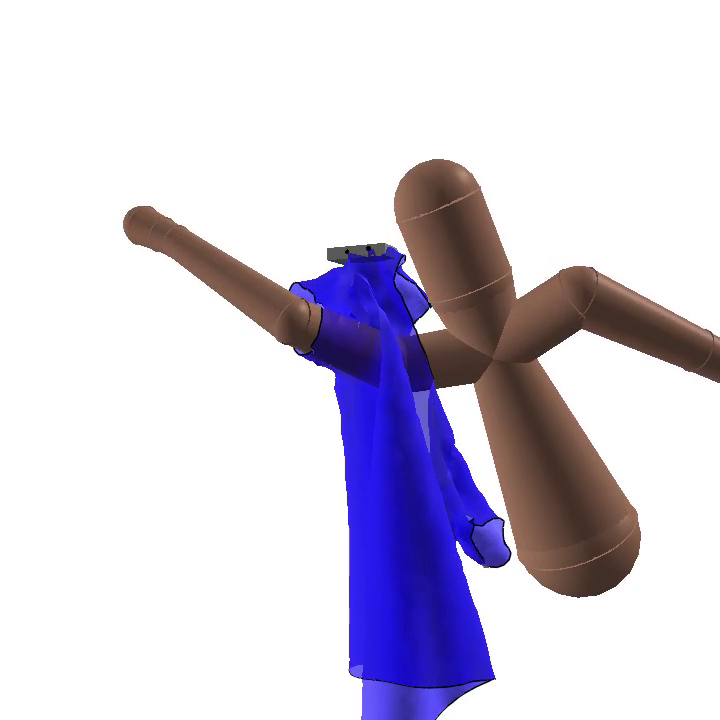} \\
\end{tabular}
\caption{Still frames of the our trained policy executing one rollout of the Side Linear Gown task.}
\label{fig:mode6figure}
\end{figure*}

We validate our approach by comparing a policy trained with our proposed combination of reward and observation terms to two baseline policies: a haptic unaware policy lacking the deformation reward term in Section \ref{deformationpenalty} and the haptic observation term in Section \ref{haptics}, and a task unaware policy that lacks the task observation feature described in Section \ref{taskvector}. These baseline policies are trained on the Fixed Gown task described in Section \ref{fixedgown} and we compare their performance with our complete policy using two metric: Arm Progress (Section \ref{progressreward}) and Max Deformation (Section \ref{deformationpenalty}). For each policy, we sample 100 random initializations and plot the mean and standard deviation of these two metrics over the rollout horizon of 400 steps. Aside from these metrics, we strongly encourage the reader to view a comparison of the motions produced by these policies in the supplemental video.

We choose to compare these policies via the progress and deformation metrics because we believe that a successful dressing policy is one that completes the task as well as possible within the constraints of minimizing damage to the garment, the robot and the human being dressed. The plots in Figure \ref{fig:baselines} and the show that while the haptic unaware baseline policy (middle) performs well on the progress metric, demonstrating that it can accomplish the task without the haptic feature, it exhibits excessive garment deformation, increasing well beyond the penalty threshold of 15. This indicates a careless dressing strategy that is unlikely to be representative of a cooperative human. However, while the task unaware baseline policy (right) successfully limits deformation, it fails to complete the task, typically hovering the end effector in the vicinity of the garment feature (sleeve). Our proposed policy (left) balances these goals, completing the task reasonably well while also achieving insignificant garment deformation. 

Figure \ref{fig:meanbars} presents a comparison of the deformation metric over the full rollout length and mean progress at the final state of each rollout. This comparison further confirms the weaknesses of each baseline while also demonstrating the potential benefits of coordinated motion between the robot and the human. The mean deformation graph shows that all policies trained with deformation penalty and haptic awareness manage similarly reasonable performance, while the average state of the haptic unaware policy exceeds the maximum deformation penalty threshold of 20. When compared to the two linear trajectory tasks, the complete policy trained on the Fixed Gown task appears to be somewhat challenging due to a lower mean progress in the final state. Additionally, Figure \ref{fig:linearpolicies} shows that both the policies that give naive robot assistance in the form of linear gripper trajectories performed well on both deformation and progress metrics. This indicates that robotic assistance can be both helpful for dressing and that trained human policies could be used to evaluate the helpfulness of particular assistive robot policies using similar metrics to those presented in this work.

\section{CONCLUSION}\label{conclusion}

We have demonstrated the creation of a model of human behavior during robotic assistance with dressing. The policy for human behavior is trained through reinforcement learning using a task-specific observed state and reward function.  The resulting policies noticeably outperform policies that omit haptic feedback or that do not make use of a task vector for directing progress.

\addtolength{\textheight}{-10cm}   





\section*{ACKNOWLEDGMENTS}

We thank Ariel Kapusta and Zackory Erickson for their help with this work.

\bibliographystyle{IEEEtran}
\bibliography{reference_bib}

\end{document}